\pdfoutput=1

\documentclass[11pt]{article}

\usepackage{acl}

\usepackage{times}
\usepackage{latexsym}
\usepackage{multicol}
\usepackage{booktabs}
\usepackage[fleqn]{amsmath}
\usepackage{multirow}
\usepackage[normalem]{ulem}
\usepackage{hyperref}
\usepackage{graphicx}
\usepackage{subfigure}
\usepackage{fixltx2e}
\usepackage{algorithm}
\usepackage{algorithmic}
\usepackage{setspace}
\usepackage{soul}
\usepackage{amsmath}
\usepackage{amsthm}
\usepackage{amssymb}
\usepackage{colortbl}
\usepackage{xcolor}
\usepackage{pifont}
\newcommand{\ours}{\textit{ConvAgent}}

\usepackage{listings}
\usepackage{xcolor}
\lstdefinelanguage{PolicyPrompt}{
  literate=
    {<think>}{{{\color{green!60!black}<think>}}}{7}
    {</think>}{{{\color{green!60!black}</think>}}}{8}
    {<search>}{{{\color{orange!80!black}<search>}}}{8}
    {</search>}{{{\color{orange!80!black}</search>}}}{9}
    {<clarify>}{{{\color{pink!60!black}<clarify>}}}{10}
    {</clarify>}{{{\color{pink!60!black}</clarify>}}}{11}
    {<information>}{{{\color{blue!80!black}<information>}}}{13}
    {</information>}{{{\color{blue!80!black}</information>}}}{14}
    {<answer>}{{{\color{red!70!black}<answer>}}}{8}
    {</answer>}{{{\color{red!70!black}</answer>}}}{9}
    {Historical Context}{{{\color{red!90!black}Historical Context}}}{18}
    {Current User Query}{{{\color{red!90!black}Current User Query}}}{19}
}

\usepackage[T1]{fontenc}

\usepackage[utf8]{inputenc}

\usepackage{microtype}

\usepackage{inconsolata}

\usepackage{graphicx}

\title{Agentic Conversational Search with Contextualized Reasoning via Reinforcement Learning}

\author{Fengran Mo$^1$\thanks{Work done during the internship in Amazon.}, Yifan Gao$^2$, Sha Li$^2$, Hansi Zeng$^{3*}$, Xin Liu$^2$, Zhaoxuan Tan$^{4*}$\\ \textbf{Xian Li}$^2$, \textbf{Jianshu Chen}$^2$, \textbf{Dakuo Wang}$^5$, \textbf{Meng Jiang}$^4$\\
$^1$University of Montreal; 
$^2$Amazon.com;
$^3$University of Massachusetts Amherst\\ 
$^4$University of Notre Dame;  $^5$Northeastern University\\ 
\texttt{fengran.mo@umontreal.ca, yifangao@amazon.com} \\
}

\begin{document}
\maketitle
\begin{abstract}
Large Language Models (LLMs) have become a popular interface for human–AI interaction, supporting information seeking and task assistance through natural, multi-turn dialogue.
To respond to users within multi-turn dialogues, the context-dependent user intent evolves across interactions, requiring contextual interpretation, query reformulation, and dynamic coordination between retrieval and generation.
Existing studies usually follow static ``rewrite, retrieve, and generate'' pipelines, which optimize different procedures separately and overlook the mixed-initiative action optimization simultaneously.
Although the recent developments in deep search agents demonstrate the effectiveness in jointly optimizing retrieval and generation via reasoning, these approaches focus on single-turn scenarios, which might lack the ability to handle multi-turn interactions.
We introduce a conversational agent that interleaves search and reasoning across turns, enabling exploratory and adaptive behaviors learned through reinforcement learning (RL) training with tailored rewards towards evolving user goals.
The experimental results across four widely used conversational benchmarks demonstrate the effectiveness of our methods by surpassing several existing strong baselines.
\end{abstract}

\section{Introduction}
Large Language Models (LLMs) have become a popular interface for human–AI interaction, supporting information seeking and task completion assistance through natural, multi-turn dialogue~\cite{gao2022neural,zamani2023conversational,mo2024survey}. To respond to users within multi-turn dialogues, the context-dependent user intent usually evolves across historical interactions, requiring contextual interpretation~\cite{jin2023instructor}, mixed-initiative actions~\cite{aliannejadi2021analysing}, and dynamic coordination between information retrieval and response generation~\cite{su2024dragin,qian2025scent,lai2025crmweaver,qi2026language}. 

To facilitate the multi-turn conversational scenarios, existing studies~\cite{roy2024learning,ye2024boosting,mo2026leveraging} usually follow static ``rewrite, retrieve, and generate'' workflows with separated models.  
Such a pipeline optimizes various procedures separately and requires different task-specific supervision signals, including rewritten queries, relevance judgments, and ground-truth answers for each turn. Besides, it might overlook the mixed-initiative action optimization, e.g., asking a clarification question at a suitable moment, simultaneously.
Aiming for a dynamic procedure, recent developments in deep search agentic models~\cite{jin2025search,zheng2025deepresearcher,zeng2026synplanresearch} achieve remarkable performance on complex information-seeking tasks by interleaving multi-round reasoning and external search. 
The flexible architecture allows it to dynamically explore external knowledge sources and iteratively refine its reasoning trajectory for open-domain question answering~\cite{zhang2024ai,zhu2025rank,zhang2025blind,su2025parametric}.
However, these advanced deep search agentic models are designed for single-turn information accessing tasks, which might lack the ability to handle multi-turn conversational scenarios~\cite{zhang2024onegen,zhu2025convsearch,mo2025conversational}. 

Under user-system multiple interactions, the user's queries are context-dependent and involve topic-switch~\cite{mao2022curriculum,mo2023learning,yoon2025ask}. Thus, it is crucial for the conversational agent to learn to leverage useful historical context to reply to the current turn via reasoning ability. 
One important aspect is to generate a stand-alone search query condition on comprehending the historical context~\cite{jang2023itercqr,mo2024chiq}, which is different from the existing single-turn deep search agent that decomposes a self-contained query.
Another crucial factor lies in optimizing the model to effectively leverage information obtained from search results through de-contextualization, enabling accurate answer generation that integrates retrieved knowledge with the model's inherent parametric knowledge~\cite{wang2024astute,su2025parametric,mo2026opendecoder}, while the existing studies on deep search lack search results optimization~\cite{jin2025search}.
Besides, the flexible interaction interface enables the system to respond in various ways, which make it possible to equip the conversational agents with mixed-initiative actions capabilities to response in suitable forms, such as asking clarification questions~\cite{zou2023users}, providing short or long form answers with dynamic decisions depending on the query type~\cite{jeong2024adaptive,mo2025towards}, and rejecting to answer when without enough evidence~\cite{wang2024not}. 
The mixed-initiative actions are useful for the system to engage with the user~\cite{aliannejadi2021analysing,mo2024history,wang2024depth}.

To address the aforementioned issues, we aim to optimize the agentic conversational search task from multiple aspects within a single agentic framework with contextualized reasoning, including history-conditional search query generation, the utilization of search results for accurate answer generation, and interactive action decision via mixed-initiative capacities.
To achieve it, we propose \ours{} to decompose the overall reward into three complementary components under a reinforcement learning (RL) training framework~\cite{shao2024deepseekmath} to interleave search and reasoning across turns, enabling exploratory and adaptive behaviors learned through tailored design.
Specifically, to help the agentic model optimize search queries through de-contextualization reasoning in conversations and learn to extract useful information from retrieved results for answer generation, we introduce a search optimization reward. This reward measures the information overlap between the retrieved content and the ground-truth answers, providing additional supervision beyond the ground-truth answer signal alone.
Besides, a mixed-initiative action reward is used to enhance the alignment with evolving user goals and facilitate the mixed-initiative capacity of the models. 
The experimental results across four widely used conversational search benchmarks demonstrate the effectiveness of our methods by surpassing several existing strong baselines with different principles under various settings. 
Detailed ablation analyses show the contributions of different components and insightful observations to improve multi-turn capacity for conversational agents.

Our contributions can be summarized as follows:

(1) We introduce an agentic framework to optimize the conversational search task from multiple aspects via contextualized reasoning, including history-conditional search query generation, the utilization of search results, and interactive action decision via mixed-initiative capacities.

(2) We propose to decompose the RL training rewards with intermediate components, including a search optimization and a mixed-initiative action reward in addition to the outcome, which enables the model with exploratory and adaptive behaviors to interleave search and reasoning across turns.

(3) We conduct experiments on four widely used conversational search benchmarks and take analyses to understand how agentic models with search and mix-initiative actions impact generation performance in conversational search.

\section{Related Work}
\noindent \textbf{Conversational Question Answering} is the task of
responding to user queries within multi-turn dialogues, with dependence on previous turns and often grounding from external knowledge~\cite{qu2020open}. 
The retrieval-augmented generation (RAG) technique~\cite{lewis2020retrieval} is a common practice to obtain external factual grounding in answers.
Previous studies in conversational search~\cite{wu2022conqrr,mo2023convgqr,abbasiantaeb2024let,lupart2025disco} aim to enhance the understanding of user intent and more complex user interactions. 
A series of approaches are proposed to improve the search results performance in conversations, including query rewriting~\cite{mao2023large,ye2023enhancing,lai2024adacqr}, conversational dense retrieval fine-tuning~\cite{lin2021contextualized,kim2022saving,mao2024chatretriever,mo2024aligning}, and data augmentation~\cite{mao2022convtrans,mo2025convmix}.
Besides, several studies~\cite{meng2023system,owoicho2023exploiting} have explored improving mixed-initiative action detection in conversational settings to provide more effective support during interactions.
However, these approaches do not directly explore how to leverage the retrieved information to improve answer generation in the era of LLMs~\cite{gao2023retrieval,mo2024survey,bhaskar2025language,zhang2025entropy,zhang2025ratt}.
Our research aims to develop an agentic conversational search model that can understand context-dependent user intent, optimize search queries, enhance mixed-initiative interaction, and reason over search results to generate accurate answers. 

\noindent \textbf{Agentic Search.}
Recent studies~\cite{zhang2024ai} focus on developing agentic LLMs for information accessing tasks via RL training.
The principle is to decompose the complex questions via integrating in-context reasoning with dynamic search tool invocation when needed.
Representative studies such as Search-r1~\cite{jin2025search}, R1-Searcher~\cite{song2025r1}, and DeepResearch~\cite{zheng2025deepresearcher} optimize RL policies for generating multiple queries through multi-turn search interactions and integrating retrieved evidence, thereby achieving strong performance on knowledge-intensive QA tasks.
However, they are limited to handling self-contained questions as one-shot querying and do not require understanding historical context.
Different from them, we aim to define an intermediate reward tailored to answer questions in conversations that directly optimizes the agentic model to utilize search results when generating rewritten queries for search in capturing user intent at each dialogue turn.
Concurrent work ChatR1~\cite{lupart2025chatr1} and ConvRecR1~\cite{zhu2025rank} attempt to extend this line of work, while their designed reward for conversational scenarios relies on the ground-truth of rewritten queries as training signals and does not consider the mixed-initiative action capability.

\section{Methodology}

\subsection{Task Formulation}
A conversational agent should be capable of performing various actions through reasoning in conversation to arrive at the final answer. 
Formally, given a dataset of user–system conversations, each composed of multiple turns, and a collection of passages $\mathcal{C}$.
At each turn $n$, the agent system receives the conversation history $\mathcal{H}_n=\{q_i,a_i\}_{i=1}^{n-1}$ and the current user query $q_n$. 
The conversational agent requires generating an answer $a_n$ to query $q_n$, leveraging context from $\mathcal{H}_n$ and grounding in external information $\mathcal{C}$. In some cases, the agent should also be capable of handling unanswerable or ambiguous queries through rejection and clarification, resulting in requiring mixed-initiative action capabilities.
We aim to develop such an agentic model via contextualized reasoning in this study.

\subsection{Rewards Modeling}
\begin{figure}[t]
\centering
\includegraphics[width=1\linewidth]{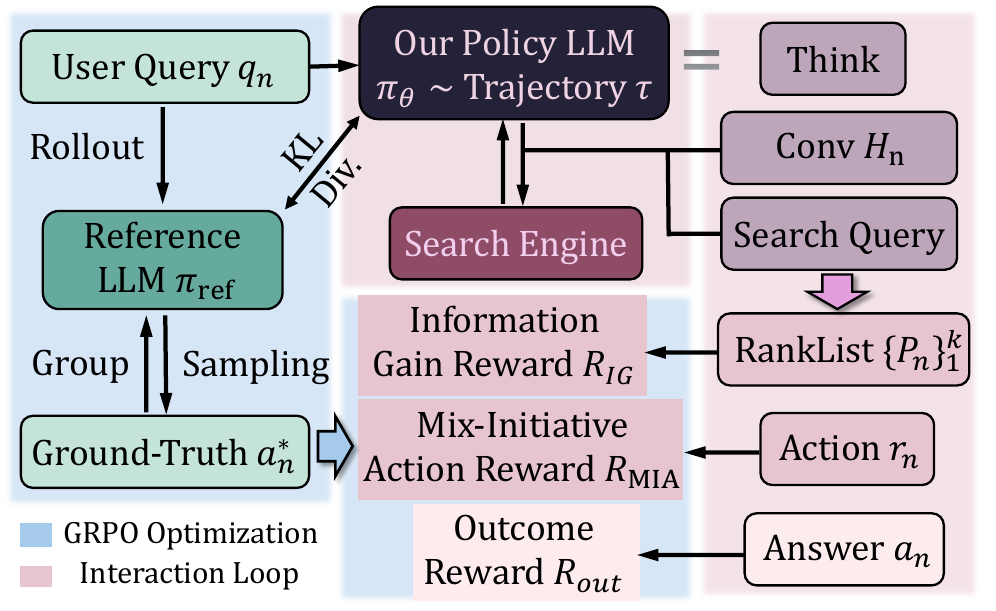}
\caption{Overview of our designed agentic conversational search framework. 
The GRPO Optimization is guided by full trajectories sampled from the policy LLM, which consists of reasoning, search, mix-initiative action, and answer steps, with rewards assigned to each signal.
The policy model interleaves the same action set while interacting with a search environment within the Interaction Loop for both training and inference.}
\label{fig: overview}
\vspace{-2ex}
\end{figure}

Reinforcement learning (RL) plays a crucial role in training agentic models by enabling self-exploration of diverse reasoning trajectories and rewarding those that are effective.
This aligns model behavior with the objectives of information seeking for users' context-dependent intents and efficient decision-making for actions in conversations. 
However, since these trajectories are self-generated and do not always yield correct final answers due to noisy conversational context, relying solely on outcome-based rewards yields sparse supervision signals, making the training process harder to optimize~\cite{chen2025towards,zhang2026starpo}.
To this end, we decompose the overall reward into three complementary components: outcome reward, search optimization reward, and mixed-initiative action reward. These rewards guide the model to reason more strategically with the conversational context towards the final answer, effectively capture user intent for search optimization, and respond precisely to various types of queries with mixed-initiative needs in conversations.
We describe the overview of our framework in Figure~\ref{fig: overview}.

\subsubsection{Outcome Generation Reward}
The outcome generation reward directly reflects the final task success.
At the end of each rollout, the predicted answer $a_n$ is extracted to evaluate by comparing with the ground-truth answer $a_n^*$ using a task-specific metric $\mathcal{S}_{\text{ans}}$ as $\mathcal{S}_{\text{ans}}(a_n, a_n^*)$, e.g., Exact Match or F1 score. The formulation of the outcome generation reward is 
\begin{equation}
    \mathcal{R}_{\text{outcome}} = \mathcal{S}_{\text{ans}}(a_n, a_n^*)
\end{equation}

\subsubsection{Information Gain Reward} 
\label{sec: Information Gain Reward}
Solely relying on the supervision signal from the ground-truth answer is not enough to enable the conversational agent to interpret user intent and establish high-quality intermediate results for external search. 
Different from previous studies~\cite{jin2025search,song2025r1}, which only generate search queries for calling the search interface, the agentic model requires optimizing the search query via de-contextualization in conversations and learn to utilize the gained information via external retrieval, i.e., the useful parts of the search results in terms of approaching the final answer. 
To this end, the information gain reward is designed to measures the information overlap between the retrieved information and the ground-truth $a_n^*$ via applying a specific metric $\mathcal{S}_{\text{Info}}$ on them as $\mathcal{S}_{\text{Info}}(\mathcal\{{P}_n\}_1^k, a_n^*)$. 
The $\mathcal\{{P}_n\}_1^k$ is the top-$k$ relevant passages obtained by an ad-hoc retriever $\mathcal{R}\text{et}$ via the generated rewritten-query $q_n^{\prime}$ for the $n$-th turn as $\mathcal\{{P}_n\}_1^k = \mathcal{R}\text{et}(q_n^{\prime})$.

Such an optimization is expected to improve the quality of the agent-rewritten search query by aligning its search results with the associated final answer without requiring the relevance judgments between query and gold passages as supervision signals.
The information gain reward $\mathcal{R}_{\text{IG}}$ is formulated as the cumulative coverage of the search results in each retrieval calling iteration. According to the length of the ground-truth answer for each query~\cite{bolotova2022non}, we use either F1-score or Accuracy as an evaluation metric to validate the information gain reward from long or short types of answers, which is defined as
$$
\begin{aligned}         & \mathcal{R}_{\text{IG}} = \mathcal{S}_{\text{Info}}(\mathcal{R}\text{et}(q_n^{\prime}), a_n^*) = \mathcal{S}_{\text{Info}}(\mathcal\{{P}_n\}_1^k, a_n^*) \\
&= 
\begin{cases}
\max_{q_n^{\prime}} \mathrm{F1}(\mathcal\{{P}_n\}_1^k, a_n^*) & \text{type$(a_n^*)$ = long} \\
\max_{q_n^{\prime}} \mathrm{Acc}(\mathcal\{{P}_n\}_1^k, a_n^*) & \text{type$(a_n^*)$ = short}
\end{cases}
\end{aligned}
$$
where the accuracy is measured by whether the ground-truth $a_n^*$ is a sub-string contained in the retrieved information $\mathcal\{{P}_n\}_1^k$ as $\mathbb{I}\big[a_n^* \subseteq \mathcal\{{P}_n\}_1^k\big]$. \\

\subsubsection{Mixed-Initiative Action Reward} 
The mixed-initiative plays an important role in conversational scenarios, where the agentic model is expected to clarify ambiguous query turns or refuse to answer when their knowledge is insufficient.
To facilitate the mixed-initiative capacity of our model, we formulate the mixed-initiative as a classification task to validate whether the agentic model can decide the necessary supportive reactions. 
Specifically, we include four types of mixed-initiative action following~\cite{wu2023inscit} in the reaction set as $\mathcal{A}\text{ct} = \{\text{answer}, \text{clarify}, \text{no answer}\}$. In addition to generating a final answer as a common case, the agentic model should generate clarification questions, return only relevant search results with the final answer, or refuse to answer to accommodate the unanswerable cases.
Then, the mixed-initiative action (MIA) reward $\mathcal{R}_{\text{MIA}}$ is designed similarly to the format reward to measure whether the specific label tokens, i.e., <clarify> and <noanswer>, are included in the generated sequence $r_n$ among the turns that are annotated with requiring specific mixed-initiative actions in the ground-truth reaction set $\mathcal{A}\text{ct}_n^{*}$ as
\begin{equation}
\mathcal{R}_{\text{MIA}} =
\begin{cases}
1, & \text{if } {r_n} \subseteq \mathcal{A}\text{ct}_n^{*},\\[6pt]
-0.5, & \text{otherwise.}
\end{cases}
\end{equation}
The negative reward assigned to incorrect mixed-initiative actions serves as a penalty that discourages the model from aggressively generating unnecessary clarifications or refusals.

Finally, the reward assigned to a rollout trajectory $\tau$ aggregates the three components as
\begin{equation}
\label{eq: final_reward}
    \mathcal{R}(\tau) = \mathcal{R}_{\text{outcome}} + 0.5 \times (\mathcal{R}_{\text{IG}} + \mathcal{R}_{\text{MIA}})
\end{equation}

\subsection{Contextualized Reasoning in Conversations with RL}
With the designed rewards, the optimization objective of our model with parameters $\theta$ is to maximize the aggregated reward $\mathcal{R}(\tau)$ given the history $\mathcal{H}$, the user query $q$ at the last turn $n$, and the search engine $\mathcal{R}\text{et}$, while minimizing the distance between the optimized and original policies as Eq.~\ref{eq: objective}.
\begin{equation}
\begin{aligned}
\mathcal{J}_\text{GRPO}&(\theta) = \mathbb{E}\Big[
\sum_{i} \min\Big(\phi_i(\theta)\mathcal{A}_i, \
\text{clip}\Big(\phi_i(\theta), 
\\& 1-\epsilon, 1+\epsilon\Big)\mathcal{A}_i\Big)
- \gamma D_{KL}(\pi_{\theta}||\pi_{ref})\Big]
\end{aligned}
\label{eq: objective}
\end{equation}
where $\phi_i(\theta)=\frac{\pi_{\theta}(\tau_i|q, \mathcal{H},\tau_{<i}, \mathcal{R}\text{et})}{\pi_{\theta_{old}}(\tau_i|q, \mathcal{H},\tau_{<i}, \mathcal{R}\text{et})}$ denotes the probability between the new and old policies and $\mathcal{A}_i=\frac{\mathcal{R}(\tau_i) - \text{mean}(\mathcal{R}(\tau))}{\text{std}(\mathcal{R}(\tau))}$ represents the estimated advantage of the $i$-th rollout trajectory within the current group calculated by our designed reward function $\mathcal{R}(\tau)$ in Eq.~\ref{eq: final_reward}. Besides, the parameter $\epsilon$ controls the clipping threshold, while $\gamma$ regulates the KL divergence penalty. 

We adopt the efficient Group Relative Policy Optimization (GRPO) algorithm \cite{shao2024deepseekmath}, which eliminates the need for explicit reward and value models.
As an alternative optimization approach, we also implement the Proximal Policy Optimization (PPO) to optimize our approach. The corresponding analysis compared to GRPO is shown in the experimental Section~\ref{sec: Ablation Studies}.

\subsection{Usage of Clarification Results}
The goal of generating necessary clarified questions in mixed-initiative actions for some ambiguous turns is to support the user in obtaining their desired answers. 
Thus, the evaluation of clarification actions should consider not only whether they are taken at appropriate moments but also how effective they are in improving downstream task performance. To this end, we design the usage of clarification results in downstream tasks. 
Specifically, in the retrieval procedure interacting with the search engine, the clarified question $q_n^c$ generated by the model is used as an expansion, concatenated with the previous rewritten query $q_n^{\prime}$ as $\mathcal{R}\text{et}(q_n^{\prime} \circ q_n^c)$ to obtain search results. For the answer generation procedure, we replace the original query $q_n$ with a clarified one $q_n^c$, which is used as a simulated question to obtain the final answer $a_n$. 

\begin{table*}[t]
\centering
\resizebox{\textwidth}{!}{
\begin{tabular}{lccccccccc}
\toprule
\multirow{2}{*}{Method} & \multirow{2}{*}{LLM} & \multicolumn{2}{c}{TopiOCQA} & \multicolumn{2}{c}{INSCIT} & \multicolumn{2}{c}{QReCC} & \multicolumn{2}{c}{CORAL}\\
\cmidrule(lr){3-4}\cmidrule(lr){5-6}\cmidrule(lr){7-8}\cmidrule(lr){9-10}
~ & ~ & F1 & EM & F1 & LLM-Jud. & F1 & LLM-Jud. & F1 & LLM-Jud.\\
\midrule
SFT & Qwen-2.5-3b & 18.2 & 14.7 & 23.7 & 52.6 & 17.0 & 50.4 & 15.2 & 42.3\\
Search-R1-3b  & Qwen-2.5-3b & 26.1 & 13.1 & 5.8 & 50.1 & 5.9 & 48.7 & 3.9 & 41.2\\
ConvSearch-R1  & Qwen-2.5-3b & 7.6 & 3.5 & 15.7 & 48.5 & - & - & - & -\\
AgenticLM-4b  & Qwen-3-4b & 29.4 & 11.2 & 21.6 & 50.8 & 7.9 & 46.3 & 22.1 & 42.6\\
ChatR1-3b  & Qwen2.5-3b & 29.4 & - & 28.2 & - & 28.0 & - & - & - \\
\midrule
\ours{}-3b (ours)  & Qwen2.5-3b & 25.2 & 14.1 & 23.5 & 51.8 & 24.1	& 56.3 & 22.4 & 43.2 \\
\midrule
\midrule
ChatQA  & LLaMA-3-8b & 18.1 & - & 25.7 & - & 23.7 & - & 20.3 & -\\
UniConv  & Mistral-2-7b & 29.6 & - &  \textbf{33.2} & - & 26.2 & - & 24.3 & -\\
EvoRAG  & Qwen2.5-7b & 26.8 & - & 26.8 & - & 24.0 & - & \underline{25.1} & - \\
REFRAG  &  LLaMA-2-7b & 28.2 & - & - & - & 17.4 & - & - & -\\
SFT & Qwen-2.5-7b & 23.6 & 17.8 & 24.5 & 53.9 & 19.1 & 54.3 & 18.8 & 43.0\\
Search-R1-7b  & Qwen2.5-7b & \underline{37.0} & \underline{20.7} & 9.1 & \underline{57.7} & 8.6 & \underline{60.3} & 3.8 & 43.0\\
AgenticLM-8b  & Qwen-3-8b & 34.6 & 15.3 & 23.2 & 52.1 & 21.1 & 49.6 & 24.1 & \underline{44.7}\\
ChatR1-7b  & Qwen2.5-7b & 30.6 & - & 27.8 & - & \textbf{31.0} & - & - & -\\
\midrule
\ours{}-7b (ours)  & Qwen2.5-7b & \textbf{40.3}$^\dagger$ & \textbf{23.3}$^\dagger$ & \underline{30.1}$^\dagger$ & \textbf{58.5}$^\dagger$ & \underline{30.4}$^\dagger$ & \textbf{64.6}$^\dagger$ & \textbf{28.9}$^\dagger$ & \textbf{46.8}$^\dagger$\\
\midrule
\multicolumn{10}{c}{For Reference - Direct Inference with Proprietary LLMs}  \\
\midrule
ChatGPT & GPT-3.5 & 25.5 & - & 22.6 & - & 22.8 & - & 26.8 & - \\
Claude & Sonnet-3.5 & 27.2 & - & 25.0 & - & 27.0 & - & 27.4 & - \\
\bottomrule
\end{tabular}}
\caption{Performance of different systems for the answer generation task. \textbf{Bold} and \underline{underline} indicates the best and the second-best results. Superscript $\dagger$ denotes significant improvements with paired t-test at $p<0.05$ over two comparable main competitors, agentic systems Search-R1 and AgenticLM.}
\label{table: Main results}
\end{table*}

\section{Experiments}

\subsection{Experimental Setup}
\noindent \textbf{Datasets.} 
We conduct the main evaluation on four widely-used conversational search datasets, including TopiOCQA~\cite{adlakha2022topiocqa}, INSCIT~\cite{wu2023inscit} for in-domain evaluation, and QReCC~\cite{anantha2021open}, and CORAL~\cite{cheng2025coral} for out-of-domain evaluation. 
More statistical details are provided in Appendix~\ref{sec: appendix_dataset}. \\

\noindent \textbf{Evaluation Metric.} 
We investigate three different capabilities of the conversational agent: search optimization with query generation, answer generation with retrieval augmentation, and action determination in conversations. 
For search evaluation, we adopt the traditional retrieval metric NDCG@3 and the token-/string–based overlap InfoGain metric to measure the overlap between retrieved passages and ground-truth answers, thereby assessing information gain \cite{qian2025scent}, which aligns with our search optimization reward.
For answer generation, we use F1-score following previous studies~\cite{lupart2025chatr1,mo2025uniconv} for fair comparison, Exact Match (EM) for the short answer dataset TopiOCQA, and LLM-as-a-Judge~\cite{gu2024survey} based on Qwen-3-4B model for the remaining towards semantic measurement.
For mixed-initiative actions, the evaluation is conducted as a classification task and then further evaluates the improvement with supportive actions.

\noindent \textbf{Baselines.}
We compare our model with various baselines, which is categorized into two types: i) Separated retriever-generator pipeline systems, including ChatQA~\cite{liu2024chatqa}, UniConv~\cite{mo2025uniconv}, REFRAG~\cite{lin2025refrag}, and EvoRAG~\cite{cheng2025evorag}, and ii) agentic systems, including Search-R1~\cite{jin2025search}, ConvSearch-R1~\cite{zhu2025convsearch}, ChatR1~\cite{lupart2025chatr1}, and the AgenticLM~\cite{yang2025qwen3} using latest Qwen-3 models as backbone with agentic capacity. 
Among them, ConvSearch-R1 and ChatR1 require the rewritten queries from original datasets or LLM-generated pseudo data as additional supervised signal during the model training.
In addition, we include a supervised fine-tuning (SFT) baseline trained with the standard next-token prediction loss for answer generation, as well as direct inference results from proprietary large language models, including ChatGPT~\cite{chatgpt} and Claude~\cite{claude}, for reference.
The detailed information about each baseline method is provided in Appendix~\ref{sec: appendix_baseline}.

\noindent \textbf{Implementation Details.}
We implement our conversational agent model based on Qwen-2.5-3b/7b~\cite{yang2025qwen3} and launch a retriever model with E5~\cite{wang2022text} based on four NVIDIA A100 80GB GPUs. The top-3 retrieved passages are used as search results for each turn among the compared baselines and our method to ensure a fair comparison.
The TopiOCQA and INSCIT are used as the training set, and evaluated on all datasets with in-domain and out-of-domain settings.
The mixed-initiative action reward is applied on the INSCIT dataset, which is the only one containing corresponding supervision signals.
The batch size of the policy model is 256, a max prompt length of 8192 tokens, and a learning rate of 1e-6. The total training steps are 150.

\begin{table}[t]
\centering
\small
\scalebox{0.87}{
\begin{tabular}{lcccc}
\toprule
Method & \small{TopiOCQA} & \small{INSCIT} & \small{QReCC} & \small{CORAL}\\
\midrule
Search-R1-3b & 22.8 & 21.4 & 35.9 & 32.9\\
AgenticLM-4b & 18.7 & 24.1 & 33.2 & 28.6 \\
ChatR1-3b & 24.1 & - & 36.4 & - \\
\midrule
\ours{}-3b & 22.7 & 24.7 & 37.2 & 31.6 \\
\midrule
\midrule
Search-R1-7b & 26.5 & 22.3 & 36.7 & 33.3 \\
AgenticLM-8b & 20.4 & 23.8 & 36.0 & 30.9 \\
ChatR1-7b & 26.7 & - & 37.0 & - \\
\midrule
\ours{}-7b  & \textbf{29.4} & \textbf{30.8} & \textbf{39.6} & \textbf{34.9}\\
\bottomrule
\end{tabular}}
\caption{Performance of the retrieval task.}
\label{table: Retrieval Results}
\vspace{-2ex}
\end{table}

\begin{table*}[t]
\centering
\scalebox{0.8}{
\begin{tabular}{lcccccccccccc}
\toprule
\multirow{2}{*}{Method} & \multicolumn{3}{c}{TopiOCQA} & \multicolumn{3}{c}{INSCIT} & \multicolumn{3}{c}{QReCC} & \multicolumn{3}{c}{CORAL}\\
\cmidrule(lr){2-4}\cmidrule(lr){5-7}\cmidrule(lr){8-10}\cmidrule(lr){11-13}
~ & N@3 & F1 & EM & N@3 & F1 & LLM-Jud. & N@3 & F1 & LLM-Jud. & N@3 & F1 & LLM-Jud.\\
\midrule
Our Full Model & 29.4 & 40.3 & 23.3 & 30.8 & 30.1 & 58.5 & 39.3 & 27.4 & 63.6 & 34.9 & 28.9 & 46.8\\
- IG Reward & 26.7 & 37.8 & 20.8 & 23.5 & 17.6 & 54.8 & 37.0 & 11.2 & 60.7 & 32.6 & 14.8 & 44.7\\
- MIA Reward & 28.1 & 39.5 & 21.2 & 29.5 & 24.3 & 60.3 & 39.6 & 30.4 & 64.6 & 33.6 & 24.1 & 46.4\\
- both & 26.5 & 37.0 & 20.7 & 22.3 & 9.1 & 50.7 & 36.7 & 8.6 & 60.3 & 33.3 & 3.8 & 43.0\\
\midrule
Replace by PPO & 26.4 & 37.4 & 21.6 & 25.7 & 29.4 & 57.5 & 37.2 & 24.7 & 63.0 & 33.8 & 20.6 & 46.5\\
\bottomrule
\end{tabular}}
\caption{Ablation studies in terms of various reward signals based on the 7b model across four datasets.}
\label{table: Ablation studies}
\end{table*}

\subsection{Main Results}
In this section, we evaluate the performance of answer generation, search results, and the correlation analysis between search results quality and answer generation performance.

\subsubsection{Answer Generation Performance}
The result of the answer generation is shown in Table~\ref{table: Main results}, where our approach consistently achieves the best or second-best performance across nearly all datasets and metrics.
Specifically, our agent model with 7B variant achieves the highest F1-scores on TopiOCQA, QReCC, and CORAL datasets by surpassing the comparable agentic systems such as AgenticLM, Search-R1, and ChatR1, and even exceeds the performance of proprietary LLMs with direct inference (ChatGPT and Claude).
Such observations demonstrate the importance of optimizing the model to learn to interact with the search engine and utilize the useful parts of search results in conversations via our designed reward signals.

Besides, the agentic systems (Search-R1, AgenticLM, ChatR1, and ours) perform better than the traditional ``retrieval-generation'' pipeline approaches (ChatQA, UniConv, and EvoRAG) with separated fine-tuning using different supervision signals, which indicates the implicit reasoning capacity is important for conversational engagement and interacting with the search engines.
Among the agentic systems, the Search-R1 does not perform well on the datasets with long-form answers due to the lack of intermediate rewards. The ChatR1 should rely on rewritten queries as additional supervision signals to capture user intent, and this might be the potential reason for its better performance on 3b variants. Nevertheless, by scaling to 7b size, our approach outperforms ChatR1, which indicates the scaling law~\cite{chung2022scaling} should still be helpful in such a scenario.

In addition, our approach also achieves better results on LLM-as-a-Judge scores, although with less significance. The scores judged by LLM might not always align with the F1 scores, which is still an open question in evaluating long-form answers.


\subsubsection{Search Results Performance}
Table~\ref{table: Retrieval Results} presents the retrieval performance across four benchmark datasets, comparing our approach with strong agentic baselines. 
Overall, our model consistently achieves the best NDCG@3 scores on all datasets for both 3B and 7B model sizes, demonstrating its effectiveness in identifying relevant information among conversational contexts. This is attributed to the search optimization rewards for generating rewritten queries with supervision signals via contextualized reasoning.
Specifically, the improvement compared to existing systems is more notable on TopiOCQA and INSCIT datasets, where the topic-switch and mixed-initiative phenomenon make the conversational retrieval more challenging but are optimized directly by our designed rewards.
These gains indicate that our conversational agent implicitly identifies the contextual relevance better via contextualized reasoning and emphasize the necessity of optimization toward adaptive user behaviors and search quality.

\subsubsection{Correlation between Search Quality and Answer Generation Performance}
The search results evaluated by traditional IR metrics (e.g., NDCG@3) might not always reflect the effectiveness for answer generation. To this end, we investigate the correlation between search quality and answer generation performance, where the search quality is measured by the information gain during the retrieval procedure in terms of approaching the final answer as described in Section~\ref{sec: Information Gain Reward}.

\begin{figure}[t]
\centering
\includegraphics[width=1\linewidth]{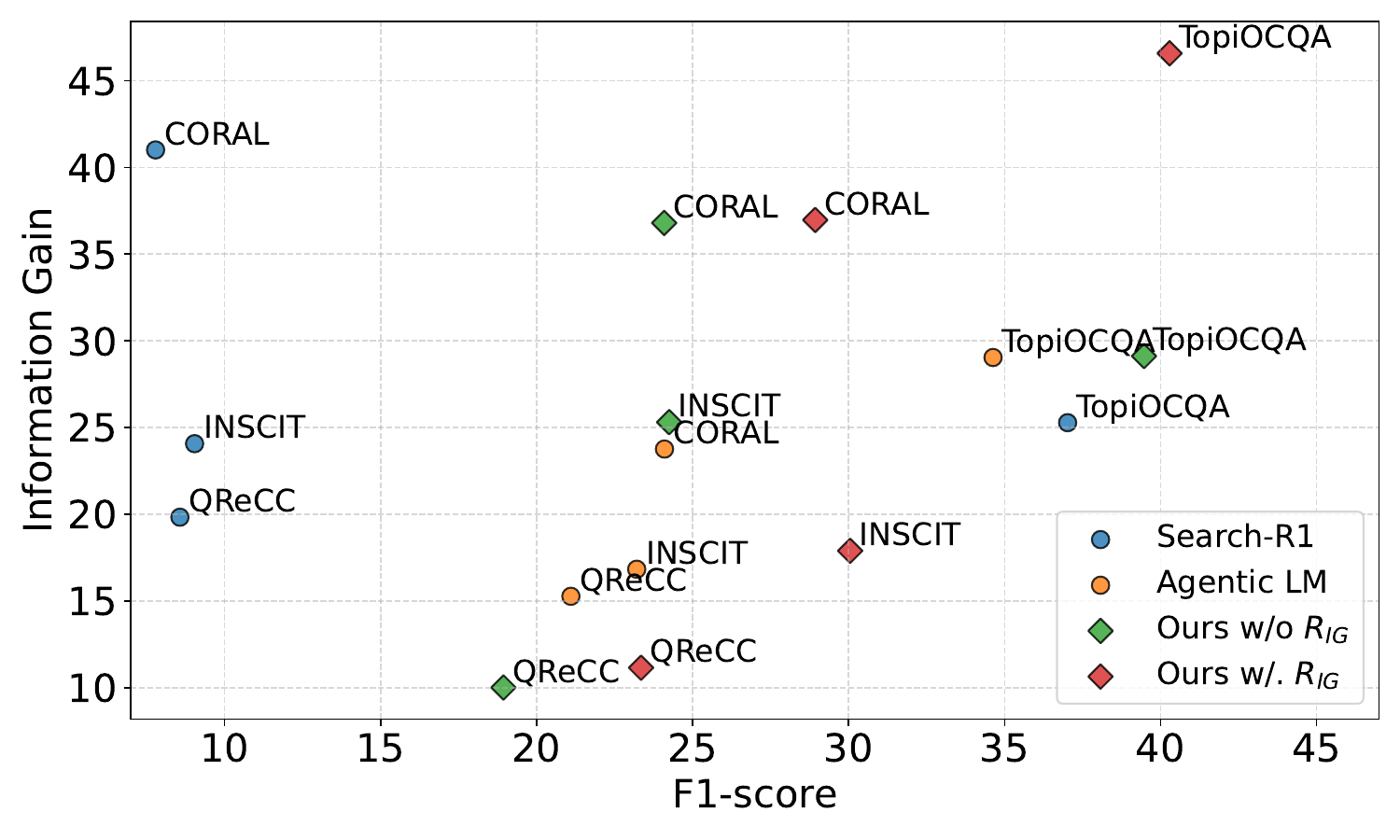}
\caption{Correlation between search quality and answer generation performance on two main competitors and our variants of w/. or w/o information gain reward.}
\label{fig: corelation}
\end{figure}

We depict the results in Figure~\ref{fig: corelation}. We observe that even with higher information gain via search results, the Search-R1 still cannot perform better in answer generation than AgenticLM and our approach with optimized capacity to utilize the retrieved knowledge from external resources. 
In other words, without specific optimization for leveraging search results, the agent models would still tend to answer the question using parametric knowledge. In contrast, our method can better integrate externally retrieved information with original knowledge via our designed information gain reward.

\begin{table*}[t]
\centering
\scalebox{0.85}{
\begin{tabular}{lcccc}
\toprule
Model & No Answer Acc. & Clarify Acc. & NDCG@3 Improve & F1 Improve\\
\midrule
\ours{} w/o $\mathcal{R}_\text{action}$ (Base-7b) & 1.42 & 3.67 & - & - \\
\ours{} w/. $\mathcal{R}_\text{action}$ \& sep. label & 11.51 & 83.19 & 3.9 & 5.8\\
\ours{} w/. $\mathcal{R}_\text{action}$ \& comb. label & 6.99 & 78.35 & 0.2 & 0.8\\
\midrule
Qwen-3-4b & 24.86 & 80.21 & -0.4 & -2.3\\
Qwen-3-8b & 20.99 & 82.00 & 0.8 & -0.2\\
Qwen-3-14b & 16.57 & 83.70 & 4.2 & 4.6\\
Qwen-3-32b & 2.21 & 81.50 & 4.8 & 4.4\\
\bottomrule
\end{tabular}}
\caption{Accuracy of mixed-initiative action and the effectiveness of clarifications on downstream tasks.}
\label{table: mixed-initiative action}
\end{table*}

\begin{figure*}[t]
\centering
\includegraphics[width=1\linewidth]{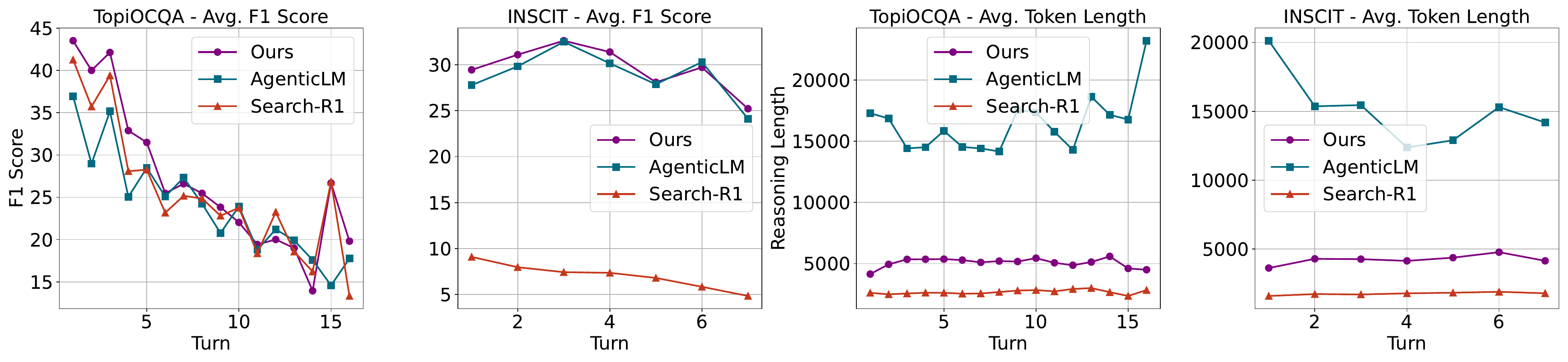}
\caption{Impact of conversational context on the reasoning effectiveness and efficiency with different models.}
\label{fig: turn-f1-length}
\end{figure*}

\subsection{Ablation Studies}
\label{sec: Ablation Studies}
The results of ablation studies are shown in Table~\ref{table: Ablation studies}, where we aim to investigate the effectiveness of various reward signals during RL training.
We find that the full model integrating both Information Gain (IG) reward and mixed-initiative action (MIA) reward achieves the best overall performance. Such results confirm these rewards as complementary benefits for reasoning when using search results for response generation in conversational scenarios, and our conjecture that solely relying on the supervision signal from ground-truth answers is not enough.
Besides, removing either reward leads to clear performance degradation. 
The IG reward contributes more than the MIA reward, while the MIA reward might not always be effective in the datasets (i.e., QReCC) that do not contain mixed-initiative phenomena.  
In addition, excluding both rewards results in a sharp decline across all datasets, and replacing the RL optimization algorithm with standard PPO yields consistently lower results, which demonstrate that the GRPO algorithm performs much better and provides more stable training in this scenario.

\subsection{Investigation of Mixed-Initiative Actions}
In this section, we investigate the performance of mixed-initiative actions.
We evaluate the accuracy of taking the associated mixed-initiative actions, whether to generate clarification (Clarify Acc.) or refuse to answer when no answers are available, at specific turns across different model capacities, including variants of our models based on ``Qwen-2.5-7b'' with various training strategies and the latest agentic models with multiple sizes.

The results are shown in Table~\ref{table: mixed-initiative action}. We observe that the non-agentic backbone model training without mixed-initiative action reward $\mathcal{R}_\text{action}$ will almost not take mixed-initiative actions. After training with our designed reward, the accuracy of determined actions improved, as well as the F1 scores of generated answers. Among them, the format of the special labels in the input template affects the final performance, where separating the labels into <clarify>, <noanswer>, and <answer> with later parsing (sep. label) can obtain better performance than combining all mixed-initiative result content in a single span (comb. label).
Moreover, the agentic models in the Qwen-3 series exhibit a clear scaling effect: larger models achieve better performance in both answer generation and clarification tasks, although they tend to respond more aggressively, even in the absence of ground-truth evidence.
Nevertheless, incorporating accurate mixed-initiative actions leads to notable improvements in both retrieval performance (NDCG@3) and answer generation (F1) compared to models without mixed-initiative optimization.

\subsection{Impact of Conversational Context}
In this analysis, we study the impact of the context (multi-turn conversations) on the reasoning capacity to leverage the historical context learned by different models. We use their per-turn answer generation performance and reasoning token length for effectiveness and efficiency evaluation.

As shown in Figure~\ref{fig: turn-f1-length}, all models drop as the conversation goes on in TopiOCQA datasets, while our model consistently performs slightly better than the others. In the INSCIT dataset, the agentic models (ours and Agentic LM) exhibit their capacity in leveraging abundant historical context in mix-initiative scenarios compared to the non-agentic one (Search-R1) and thus actually increase in longer conversations.
However, the agentic LM cannot perform well in terms of efficiency compared to Search-R1. With the intermediate rewards, our agentic model achieves a better effectiveness-efficiency trade-off, with better performance than Search-R1 and one-third of the reasoning token cost of Agentic LM.

\section{Conclusion}
In this paper, we present a contextualized reasoning RL framework to improve multi-turn capacity for conversational agents, which aims to address the scenarios of user intent evolving across turns and must be inferred from context.
We interleave search and reasoning across turns in conversations, enabling exploratory and adaptive behaviors learned through decomposed rewards by guiding the model to reason conversational context towards the final answer, effectively capture user intent for search optimization, and respond precisely to various types of queries.
The experimental results across four widely used conversational benchmarks demonstrate the effectiveness of our methods by surpassing several existing strong baselines.

\section*{Limitations}
Despite our comprehensive studies, some potential limitations can be addressed in the future:

\noindent \textbf{Efficiency.} While the RL training facilitates the multi-turn capacity of the agentic conversational search model via contextualized reasoning, the additional computational cost is added for both the training and inference phases. Thus, efficiency optimization is required. 

\noindent \textbf{Broader Experimental Configuration.} Although our approach could generalize to other backbone models, we did not test on more types and sizes of LLMs. Besides, we only leverage the fixed hyper-parameters for reward modeling, where the exploration within broader experimental configurations could lead to better performance.

In addition, how to design better intermediate reward to facilitate the utilization of search results for better response generation quality is still an open question for further exploration.



\bibliography{custom}

@article{gao2023retrieval,
  title={Retrieval-augmented generation for large language models: A survey},
  author={Gao, Yunfan and Xiong, Yun and Gao, Xinyu and Jia, Kangxiang and Pan, Jinliu and Bi, Yuxi and Dai, Yi and Sun, Jiawei and Wang, Haofen},
  journal={arXiv preprint arXiv:2312.10997},
  year={2023}
}

@inproceedings{abbasiantaeb2024let,
  title={Let the llms talk: Simulating human-to-human conversational qa via zero-shot llm-to-llm interactions},
  author={Abbasiantaeb, Zahra and Yuan, Yifei and Kanoulas, Evangelos and Aliannejadi, Mohammad},
  booktitle={Proceedings of the 17th ACM International Conference on Web Search and Data Mining},
  pages={8--17},
  year={2024}
}

@article{chen2025towards,
  title={Towards reasoning era: A survey of long chain-of-thought for reasoning large language models},
  author={Chen, Qiguang and Qin, Libo and Liu, Jinhao and Peng, Dengyun and Guan, Jiannan and Wang, Peng and Hu, Mengkang and Zhou, Yuhang and Gao, Te and Che, Wanxiang},
  journal={arXiv preprint arXiv:2503.09567},
  year={2025}
}

@inproceedings{cheng2025coral,
  title={CORAL: Benchmarking Multi-turn Conversational Retrieval-Augmented Generation},
  author={Cheng, Yiruo and Mao, Kelong and Zhao, Ziliang and Dong, Guanting and Qian, Hongjin and Wu, Yongkang and Sakai, Tetsuya and Wen, Ji-Rong and Dou, Zhicheng},
  booktitle={Findings of the Association for Computational Linguistics: NAACL 2025},
  pages={1308--1330},
  year={2025}
}

@article{lai2025crmweaver,
  title={Crmweaver: Building powerful business agent via agentic rl and shared memories},
  author={Lai, Yilong and Yang, Yipin and Wu, Jialong and Mo, Fengran and Wang, Zhenglin and Liang, Ting and Lin, Jianguo and Yang, Keping},
  journal={arXiv preprint arXiv:2510.25333},
  year={2025}
}

@inproceedings{jeong2024adaptive,
  title={Adaptive-rag: Learning to adapt retrieval-augmented large language models through question complexity},
  author={Jeong, Soyeong and Baek, Jinheon and Cho, Sukmin and Hwang, Sung Ju and Park, Jong C},
  booktitle={Proceedings of the 2024 Conference of the North American Chapter of the Association for Computational Linguistics: Human Language Technologies (Volume 1: Long Papers)},
  pages={7036--7050},
  year={2024}
}

@inproceedings{mo2024aligning,
  title={Aligning query representation with rewritten query and relevance judgments in conversational search},
  author={Mo, Fengran and Qu, Chen and Mao, Kelong and Wu, Yihong and Su, Zhan and Huang, Kaiyu and Nie, Jian-Yun},
  booktitle={Proceedings of the 33rd ACM International Conference on Information and Knowledge Management},
  pages={1700--1710},
  year={2024}
}

@inproceedings{mo2024chiq,
  title={CHIQ: Contextual History Enhancement for Improving Query Rewriting in Conversational Search},
  author={Mo, Fengran and Ghaddar, Abbas and Mao, Kelong and Rezagholizadeh, Mehdi and Chen, Boxing and Liu, Qun and Nie, Jian-Yun},
  booktitle={Proceedings of the 2024 Conference on Empirical Methods in Natural Language Processing},
  pages={2253--2268},
  year={2024}
}

@inproceedings{cheng2025evorag,
  title={Evolving Graph-Based Context Modeling for Multi-Turn Conversational Retrieval-Augmented Generation},
  author={Cheng, Yiruo and Qian, Hongjin and Mo, Fengran and Wu, Yongkang and Li, Zhonghua and Ye, Qi and Wen, Ji-Rong and Dou, Zhicheng},
  booktitle={Proceedings of the 34th ACM International Conference on Information and Knowledge Management},
  year={2025}
}

@article{jin2025search,
  title={Search-r1: Training llms to reason and leverage search engines with reinforcement learning},
  author={Jin, Bowen and Zeng, Hansi and Yue, Zhenrui and Yoon, Jinsung and Arik, Sercan and Wang, Dong and Zamani, Hamed and Han, Jiawei},
  journal={arXiv preprint arXiv:2503.09516},
  year={2025}
}

@article{zhu2025convsearch,
  title={ConvSearch-R1: Enhancing Query Reformulation for Conversational Search with Reasoning via Reinforcement Learning},
  author={Zhu, Changtai and Wang, Siyin and Feng, Ruijun and Song, Kai and Qiu, Xipeng},
  journal={arXiv preprint arXiv:2505.15776},
  year={2025}
}

@article{lupart2025chatr1,
  title={ChatR1: Reinforcement Learning for Conversational Reasoning and Retrieval Augmented Question Answering},
  author={Lupart, Simon and Aliannejadi, Mohammad and Kanoulas, Evangelos},
  journal={arXiv preprint arXiv:2510.13312},
  year={2025}
}

@inproceedings{zhang2025ratt,
  title={Ratt: A thought structure for coherent and correct llm reasoning},
  author={Zhang, Jinghan and Wang, Xiting and Ren, Weijieying and Jiang, Lu and Wang, Dongjie and Liu, Kunpeng},
  booktitle={Proceedings of the AAAI Conference on Artificial Intelligence},
  volume={39},
  number={25},
  pages={26733--26741},
  year={2025}
}

@misc{chatgpt,
  author       = {OpenAI},
  title        = {OpenAI: Introducing ChatGPT},
  year         = {2022},
  month        = {11},
  day          = {30},
  url          = {https://openai.com/index/chatgpt/}
}

@misc{claude,
  author       = {Anthropic},
  title        = {Introducing Claude},
  year         = {2023},
  month        = {3},
  day          = {14},
  url          = {https://www.anthropic.com/news/introducing-claude}
}

@article{gu2024survey,
  title={A survey on llm-as-a-judge},
  author={Gu, Jiawei and Jiang, Xuhui and Shi, Zhichao and Tan, Hexiang and Zhai, Xuehao and Xu, Chengjin and Li, Wei and Shen, Yinghan and Ma, Shengjie and Liu, Honghao and others},
  journal={arXiv preprint arXiv:2411.15594},
  year={2024}
}

@article{yang2025qwen3,
  title={Qwen3 technical report},
  author={Yang, An and Li, Anfeng and Yang, Baosong and Zhang, Beichen and Hui, Binyuan and Zheng, Bo and Yu, Bowen and Gao, Chang and Huang, Chengen and Lv, Chenxu and others},
  journal={arXiv preprint arXiv:2505.09388},
  year={2025}
}

@inproceedings{mo2025towards,
  title={Towards Adaptive Personalized Conversational Information Retrieval},
  author={Mo, Fengran and Hui, Yuchen and Tian, Yuxing and Tan, Zhaoxuan and Meng, Chuan and Su, Zhan and Huang, Kaiyu and Nie, Jian-Yun},
  booktitle={Proceedings of the 34th ACM International Conference on Information and Knowledge Management},
  pages={2137--2147},
  year={2025}
}

@article{mo2026opendecoder,
  title={OpenDecoder: Open Large Language Model Decoding to Incorporate Document Quality in RAG},
  author={Mo, Fengran and Su, Zhan and Hui, Yuchen and Zhang, Jinghan and Sun, Jia Ao and Liu, Zheyuan and Zhang, Chao and Sakai, Tetsuya and Nie, Jian-Yun},
  journal={arXiv preprint arXiv:2601.09028},
  year={2026}
}

@misc{zhang2025blind,
      title={Blind Spot Navigation in Large Language Model Reasoning with Thought Space Explorer}, 
      author={Jinghan Zhang and Fengran Mo and Tharindu Cyril Weerasooriya and Xinyue Ye and Dongjie Wang and Yanjie Fu and Kunpeng Liu},
      year={2025},
      eprint={2410.24155},
      archivePrefix={arXiv},
      primaryClass={cs.CL},
      url={https://arxiv.org/abs/2410.24155}, 
}

@article{zhang2025entropy,
  title={Entropy-based exploration conduction for multi-step reasoning},
  author={Zhang, Jinghan and Wang, Xiting and Mo, Fengran and Zhou, Yeyang and Gao, Wanfu and Liu, Kunpeng},
  journal={arXiv preprint arXiv:2503.15848},
  year={2025}
}

@inproceedings{mao2024chatretriever,
  title={ChatRetriever: Adapting Large Language Models for Generalized and Robust Conversational Dense Retrieval},
  author={Mao, Kelong and Deng, Chenlong and Chen, Haonan and Mo, Fengran and Liu, Zheng and Sakai, Tetsuya and Dou, Zhicheng},
  booktitle={Proceedings of the 2024 Conference on Empirical Methods in Natural Language Processing},
  pages={1227--1240},
  year={2024}
}

@inproceedings{mo2023learning,
  title={Learning to Relate to Previous Turns in Conversational Search},
  author={Mo, Fengran and Nie, Jian-Yun and Huang, Kaiyu and Mao, Kelong and Zhu, Yutao and Li, Peng and Liu, Yang},
  booktitle={29th ACM SIGKDD Conference On Knowledge Discover and Data Mining (SIGKDD)},
  year={2023}
}

@inproceedings{mo2023convgqr,
  title={ConvGQR: Generative Query Reformulation for Conversational Search},
  author={Mo, Fengran and Mao, Kelong and Zhu, Yutao and Wu, Yihong and Huang, Kaiyu and Nie, Jian-Yun},
  booktitle={Proceedings of the 61st Annual Meeting of the Association for Computational Linguistics (Volume 1: Long Papers)},
  pages={4998--5012},
  year={2023}
}

@inproceedings{mo2024history,
  title={History-Aware Conversational Dense Retrieval},
  author={Mo, Fengran and Qu, Chen and Mao, Kelong and Zhu, Tianyu and Su, Zhan and Huang, Kaiyu and Nie, Jian-Yun},
  booktitle={Findings of the Association for Computational Linguistics: ACL 2024},
  pages={13366--13378},
  year={2024}
}

@inproceedings{mo2025conversational,
  title={Conversational Search: From Fundamentals to Frontiers in the LLM Era},
  author={Mo, Fengran and Meng, Chuan and Aliannejadi, Mohammad and Nie, Jian-Yun},
  booktitle={Proceedings of the 48th International ACM SIGIR Conference on Research and Development in Information Retrieval},
  pages={},
  year={2025}
}

@article{su2025parametric,
  title={Parametric retrieval-augmented generation using latent routing of lora adapters},
  author={Su, Zhan and Mo, Fengran and Zhang, Jinghan and Hui, Yuchen and Sun, Jiaao and Nie, Jian-yun},
  journal={arXiv preprint arXiv:2511.17044},
  year={2025}
}

@article{zhang2026starpo,
      title={StaRPO: Stability-Augmented Reinforcement Policy Optimization}, 
      author={Jinghan Zhang and Fengran Mo and Tharindu Cyril Weerasooriya and Ruimin Dai and Xiaoyan Han and Yanjie Fu and Dakuo Wang and Kunpeng Liu},
      year={2026},
      journal={arXiv preprint arXiv:2604.08905},
}

@inproceedings{su2024dragin,
  title={Dragin: Dynamic retrieval augmented generation based on the real-time information needs of large language models},
  author={Su, Weihang and Tang, Yichen and Ai, Qingyao and Wu, Zhijing and Liu, Yiqun},
  booktitle={Proceedings of the 62nd Annual Meeting of the Association for Computational Linguistics (Volume 1: Long Papers)},
  pages={12991--13013},
  year={2024}
}

@article{qi2026language,
  title={Language-Coupled Reinforcement Learning for Multilingual Retrieval-Augmented Generation},
  author={Qi, Rui and Mo, Fengran and Chen, Yufeng and Zhang, Xue and Wang, Shuo and Li, Hongliang and Xu, Jinan and Jiang, Meng and Nie, Jian-Yun and Huang, Kaiyu},
  journal={arXiv preprint arXiv:2601.14896},
  year={2026}
}

@article{zeng2026synplanresearch,
  title={SynPlanResearch-R1: Encouraging Tool Exploration for Deep Research with Synthetic Plans},
  author={Zeng, Hansi and Li, Zoey and Gao, Yifan and Zhang, Chenwei and Pan, Xiaoman and Yang, Tao and Mo, Fengran and Lin, Jiacheng and Li, Xian and Shang, Jingbo},
  journal={arXiv preprint arXiv:2603.07853},
  year={2026}
}

@article{mo2024survey,
  title={A survey of conversational search},
  author={Mo, Fengran and Mao, Kelong and Zhao, Ziliang and Qian, Hongjin and Chen, Haonan and Cheng, Yiruo and Li, Xiaoxi and Zhu, Yutao and Dou, Zhicheng and Nie, Jian-Yun},
  journal={arXiv preprint arXiv:2410.15576},
  year={2024}
}

@inproceedings{zhang2024onegen,
  title={OneGen: Efficient One-Pass Unified Generation and Retrieval for LLMs},
  author={Zhang, Jintian and Peng, Cheng and Sun, Mengshu and Chen, Xiang and Liang, Lei and Zhang, Zhiqiang and Zhou, Jun and Chen, Huajun and Zhang, Ningyu},
  booktitle={Findings of the Association for Computational Linguistics: EMNLP 2024},
  pages={4088--4119},
  year={2024}
}

@article{chung2022scaling,
  title={Scaling instruction-finetuned language models},
  author={Chung, Hyung Won and Hou, Le and Longpre, Shayne and Zoph, Barret and Tay, Yi and Fedus, William and Li, Yunxuan and Wang, Xuezhi and Dehghani, Mostafa and Brahma, Siddhartha and others},
  journal={arXiv preprint arXiv:2210.11416},
  year={2022}
}

@article{zamani2023conversational,
  title={Conversational information seeking},
  author={Zamani, Hamed and Trippas, Johanne R and Dalton, Jeff and Radlinski, Filip and others},
  journal={Foundations and Trends{\textregistered} in Information Retrieval},
  volume={17},
  number={3-4},
  pages={244--456},
  year={2023},
  publisher={Now Publishers, Inc.}
}

@inproceedings{anantha2021open,
  title={Open-Domain Question Answering Goes Conversational via Question Rewriting},
  author={Anantha, Raviteja and Vakulenko, Svitlana and Tu, Zhucheng and Longpre, Shayne and Pulman, Stephen and Chappidi, Srinivas},
  booktitle={Proceedings of the 2021 Conference of the North American Chapter of the Association for Computational Linguistics: Human Language Technologies},
  pages={520--534},
  year={2021}
}

@article{gao2022neural,
  title={Neural approaches to conversational information retrieval},
  author={Gao, Jianfeng and Xiong, Chenyan and Bennett, Paul and Craswell, Nick},
  journal={arXiv preprint arXiv:2201.05176},
  year={2022}
}

@article{adlakha2022topiocqa,
  title={TopiOCQA: Open-domain Conversational Question Answering with Topic Switching},
  author={Adlakha, Vaibhav and Dhuliawala, Shehzaad and Suleman, Kaheer and de Vries, Harm and Reddy, Siva},
  journal={Transactions of the Association for Computational Linguistics},
  volume={10},
  pages={468--483},
  year={2022},
  publisher={MIT Press}
}

@inproceedings{aliannejadi2021analysing,
  title={Analysing mixed initiatives and search strategies during conversational search},
  author={Aliannejadi, Mohammad and Azzopardi, Leif and Zamani, Hamed and Kanoulas, Evangelos and Thomas, Paul and Craswell, Nick},
  booktitle={Proceedings of the 30th ACM International Conference on Information \& Knowledge Management},
  pages={16--26},
  year={2021}
}

@article{owoicho2023exploiting,
  title={Exploiting Simulated User Feedback for Conversational Search: Ranking, Rewriting, and Beyond},
  author={Owoicho, Paul and Sekuli{\'c}, Ivan and Aliannejadi, Mohammad and Dalton, Jeffrey and Crestani, Fabio},
  journal={arXiv preprint arXiv:2304.13874},
  year={2023}
}

@article{song2025r1,
  title={R1-searcher: Incentivizing the search capability in llms via reinforcement learning},
  author={Song, Huatong and Jiang, Jinhao and Min, Yingqian and Chen, Jie and Chen, Zhipeng and Zhao, Wayne Xin and Fang, Lei and Wen, Ji-Rong},
  journal={arXiv preprint arXiv:2503.05592},
  year={2025}
}

@inproceedings{zhang2024ai,
  title={AI Agent for Information Retrieval: Generating and Ranking},
  author={Zhang, Yongfeng and Liu, Zhiwei and Wen, Qingsong and Pang, Linsey and Liu, Wei and Yu, Philip S},
  booktitle={Proceedings of the 33rd ACM International Conference on Information and Knowledge Management},
  pages={5605--5607},
  year={2024}
}

@article{zhu2025rank,
  title={Rank-GRPO: Training LLM-based Conversational Recommender Systems with Reinforcement Learning},
  author={Zhu, Yaochen and Steck, Harald and Liang, Dawen and He, Yinhan and Ostuni, Vito and Li, Jundong and Kallus, Nathan},
  journal={arXiv preprint arXiv:2510.20150},
  year={2025}
}

@article{wang2022text,
  title={Text embeddings by weakly-supervised contrastive pre-training},
  author={Wang, Liang and Yang, Nan and Huang, Xiaolong and Jiao, Binxing and Yang, Linjun and Jiang, Daxin and Majumder, Rangan and Wei, Furu},
  journal={arXiv preprint arXiv:2212.03533},
  year={2022}
}

@article{zheng2025deepresearcher,
  title={Deepresearcher: Scaling deep research via reinforcement learning in real-world environments},
  author={Zheng, Yuxiang and Fu, Dayuan and Hu, Xiangkun and Cai, Xiaojie and Ye, Lyumanshan and Lu, Pengrui and Liu, Pengfei},
  journal={arXiv preprint arXiv:2504.03160},
  year={2025}
}

@inproceedings{mao2022curriculum,
  title={Curriculum Contrastive Context Denoising for Few-shot Conversational Dense Retrieval},
  author={Mao, Kelong and Dou, Zhicheng and Qian, Hongjin},
  booktitle={Proceedings of the 45th International ACM SIGIR Conference on Research and Development in Information Retrieval},
  pages={176--186},
  year={2022}
}

@inproceedings{wu2022conqrr,
  title={CONQRR: Conversational Query Rewriting for Retrieval with Reinforcement Learning},
  author={Wu, Zeqiu and Luan, Yi and Rashkin, Hannah and Reitter, David and Tomar, Gaurav Singh},
  booktitle={Proceedings of the 2022 Conference on Empirical Methods in Natural Language Processing (EMNLP)},
  year={2022}
}

@inproceedings{lin2021contextualized,
  title={Contextualized Query Embeddings for Conversational Search},
  author={Lin, Sheng-Chieh and Yang, Jheng-Hong and Lin, Jimmy},
  booktitle={Proceedings of the 2021 Conference on Empirical Methods in Natural Language Processing},
  pages={1004--1015},
  year={2021}
}

@inproceedings{qu2020open,
  title={Open-retrieval conversational question answering},
  author={Qu, Chen and Yang, Liu and Chen, Cen and Qiu, Minghui and Croft, W Bruce and Iyyer, Mohit},
  booktitle={Proceedings of the 43rd International ACM SIGIR conference on research and development in Information Retrieval},
  pages={539--548},
  year={2020}
}

@inproceedings{kim2022saving,
    title={Saving dense retriever from shortcut dependency in conversational search},
    author={Kim, Sungdong and Kim, Gangwoo},
    booktitle = "Proceedings of the 2022 Conference on Empirical Methods in Natural Language Processing",
    month = dec,
    year = "2022",
    publisher = "Association for Computational Linguistics",
    pages = "10278--10287",
}

@inproceedings{lai2024adacqr,
  title={AdaCQR: Enhancing Query Reformulation for Conversational Search via Sparse and Dense Retrieval Alignment},
  author={Lai, Yilong and Wu, Jialong and Zhang, Congzhi and Sun, Haowen and Zhou, Deyu},
  booktitle={COLING},
  year={2024}
}

@inproceedings{lupart2025disco,
  title={DiSCo Meets LLMs: A Unified Approach for Sparse Retrieval and Contextual Distillation in Conversational Search},
  author={Lupart, Simon and Aliannejadi, Mohammad and Kanoulas, Evangelos},
  booktitle={SIGIR},
  year={2025}
}

@article{mo2025convmix,
  title={Convmix: A mixed-criteria data augmentation framework for conversational dense retrieval},
  author={Mo, Fengran and Zhang, Jinghan and Hui, Yuchen and Sun, Jia Ao and Xu, Zhichao and Su, Zhan and Nie, Jian-Yun},
  journal={arXiv preprint arXiv:2508.04001},
  year={2025}
}

@inproceedings{mao2022convtrans,
  title={ConvTrans: Transforming Web Search Sessions for Conversational Dense Retrieval},
  author={Mao, Kelong and Dou, Zhicheng and Qian, Hongjin and Mo, Fengran and Cheng, Xiaohua and Cao, Zhao},
  booktitle={Proceedings of the 2022 Conference on Empirical Methods in Natural Language Processing},
  pages={2935--2946},
  year={2022}
}

@article{jang2023itercqr,
  title={IterCQR: Iterative Conversational Query Reformulation without Human Supervision},
  author={Jang, Yunah and Lee, Kang-il and Bae, Hyunkyung and Won, Seungpil and Lee, Hwanhee and Jung, Kyomin},
  journal={arXiv preprint arXiv:2311.09820},
  year={2023}
}

@inproceedings{yoon2025ask,
  title={Ask Optimal Questions: Aligning Large Language Models with Retriever’s Preference in Conversation},
  author={Yoon, Chanwoong and Kim, Gangwoo and Jeon, Byeongguk and Kim, Sungdong and Jo, Yohan and Kang, Jaewoo},
  booktitle={Findings of the Association for Computational Linguistics: NAACL 2025},
  pages={5899--5921},
  year={2025}
}

@article{shao2024deepseekmath,
  title={Deepseekmath: Pushing the limits of mathematical reasoning in open language models},
  author={Shao, Zhihong and Wang, Peiyi and Zhu, Qihao and Xu, Runxin and Song, Junxiao and Bi, Xiao and Zhang, Haowei and Zhang, Mingchuan and Li, YK and Wu, Yang and others},
  journal={arXiv preprint arXiv:2402.03300},
  year={2024}
}

@inproceedings{wang2024depth,
  title={An in-depth investigation of user response simulation for conversational search},
  author={Wang, Zhenduo and Xu, Zhichao and Srikumar, Vivek and Ai, Qingyao},
  booktitle={Proceedings of the ACM Web Conference 2024},
  pages={1407--1418},
  year={2024}
}

@article{qian2025scent,
  title={Scent of Knowledge: Optimizing Search-Enhanced Reasoning with Information Foraging},
  author={Qian, Hongjin and Liu, Zheng},
  journal={arXiv preprint arXiv:2505.09316},
  year={2025}
}

@inproceedings{wang2024not,
  title={Do-not-answer: Evaluating safeguards in LLMs},
  author={Wang, Yuxia and Li, Haonan and Han, Xudong and Nakov, Preslav and Baldwin, Timothy},
  booktitle={Findings of the Association for Computational Linguistics: EACL 2024},
  pages={896--911},
  year={2024}
}

@article{zou2023users,
  title={Users meet clarifying questions: Toward a better understanding of user interactions for search clarification},
  author={Zou, Jie and Aliannejadi, Mohammad and Kanoulas, Evangelos and Pera, Maria Soledad and Liu, Yiqun},
  journal={ACM Transactions on Information Systems},
  volume={41},
  number={1},
  pages={1--25},
  year={2023},
  publisher={ACM New York, NY}
}

@article{wang2024astute,
  title={Astute rag: Overcoming imperfect retrieval augmentation and knowledge conflicts for large language models},
  author={Wang, Fei and Wan, Xingchen and Sun, Ruoxi and Chen, Jiefeng and Ar{\i}k, Sercan {\"O}},
  journal={arXiv preprint arXiv:2410.07176},
  year={2024}
}

@inproceedings{jin2023instructor,
  title={InstructoR: Instructing Unsupervised Conversational Dense Retrieval with Large Language Models},
  author={Jin, Zhuoran and Cao, Pengfei and Chen, Yubo and Liu, Kang and Zhao, Jun},
  booktitle={Findings of the Association for Computational Linguistics: EMNLP 2023},
  pages={6649--6675},
  year={2023}
}

@inproceedings{ye2023enhancing,
  title={Enhancing Conversational Search: Large Language Model-Aided Informative Query Rewriting},
  author={Ye, Fanghua and Fang, Meng and Li, Shenghui and Yilmaz, Emine},
  booktitle={Findings of the Association for Computational Linguistics: EMNLP 2023},
  pages={5985--6006},
  year={2023}
}

@inproceedings{bolotova2022non,
  title={A non-factoid question-answering taxonomy},
  author={Bolotova, Valeriia and Blinov, Vladislav and Scholer, Falk and Croft, W Bruce and Sanderson, Mark},
  booktitle={Proceedings of the 45th International ACM SIGIR Conference on Research and Development in Information Retrieval},
  pages={1196--1207},
  year={2022}
}

@article{bhaskar2025language,
  title={Language models that think, chat better},
  author={Bhaskar, Adithya and Ye, Xi and Chen, Danqi},
  journal={arXiv preprint arXiv:2509.20357},
  year={2025}
}

@inproceedings{mao2023large,
  title={Large Language Models Know Your Contextual Search Intent: A Prompting Framework for Conversational Search},
  author={Mao, Kelong and Dou, Zhicheng and Chen, Haonan and Mo, Fengran and Qian, Hongjin},
  booktitle={Findings of the Association for Computational Linguistics: EMNLP 2023},
  year={2023}
}

@inproceedings{meng2023system,
  title={System Initiative Prediction for Multi-turn Conversational Information Seeking},
  author={Meng, Chuan and Aliannejadi, Mohammad and de Rijke, Maarten},
  booktitle={Proceedings of the 32nd ACM International Conference on Information and Knowledge Management},
  pages={1807--1817},
  year={2023}
}

@inproceedings{roy2024learning,
  title={Learning When to Retrieve, What to Rewrite, and How to Respond in Conversational QA},
  author={Roy, Nirmal and Ribeiro, Leonardo and Blloshmi, Rexhina and Small, Kevin},
  booktitle={Findings of the Association for Computational Linguistics: EMNLP 2024},
  pages={10604--10625},
  year={2024}
}

@article{lewis2020retrieval,
  title={Retrieval-augmented generation for knowledge-intensive nlp tasks},
  author={Lewis, Patrick and Perez, Ethan and Piktus, Aleksandra and Petroni, Fabio and Karpukhin, Vladimir and Goyal, Naman and K{\"u}ttler, Heinrich and Lewis, Mike and Yih, Wen-tau and Rockt{\"a}schel, Tim and others},
  journal={Advances in Neural Information Processing Systems},
  volume={33},
  pages={9459--9474},
  year={2020}
}

@article{wu2023inscit,
  title={INSCIT: Information-Seeking Conversations with Mixed-Initiative Interactions},
  author={Wu, Zeqiu and Parish, Ryu and Cheng, Hao and Min, Sewon and Ammanabrolu, Prithviraj and Ostendorf, Mari and Hajishirzi, Hannaneh},
  journal={Transactions of the Association for Computational Linguistics},
  volume={11},
  pages={453--468},
  year={2023},
  publisher={MIT Press One Broadway, 12th Floor, Cambridge, Massachusetts 02142, USA~…}
}

@article{liu2024chatqa,
  title={Chatqa: Surpassing gpt-4 on conversational qa and rag},
  author={Liu, Zihan and Ping, Wei and Roy, Rajarshi and Xu, Peng and Lee, Chankyu and Shoeybi, Mohammad and Catanzaro, Bryan},
  journal={Advances in Neural Information Processing Systems},
  volume={37},
  pages={15416--15459},
  year={2024}
}

@inproceedings{ye2024boosting,
  title={Boosting conversational question answering with fine-grained retrieval-augmentation and self-check},
  author={Ye, Linhao and Lei, Zhikai and Yin, Jianghao and Chen, Qin and Zhou, Jie and He, Liang},
  booktitle={Proceedings of the 47th International ACM SIGIR Conference on Research and Development in Information Retrieval},
  pages={2301--2305},
  year={2024}
}

@article{mo2026leveraging,
  title={Leveraging historical information to boost retrieval-augmented generation in conversations},
  author={Mo, Fengran and Gao, Yifan and Wu, Zhuofeng and Liu, Xin and Chen, Pei and Li, Zheng and Wang, Zhengyang and Li, Xian and Jiang, Meng and Nie, Jian-Yun},
  journal={Information Processing \& Management},
  volume={63},
  number={2},
  pages={104449},
  year={2026},
  publisher={Elsevier}
}

@inproceedings{mo2025uniconv,
  title={UniConv: Unifying Retrieval and Response Generation for Large Language Models in Conversations},
  author={Mo, Fengran and Gao, Yifan and Meng, Chuan and Liu, Xin and Wu, Zhuofeng and Mao, Kelong and Wang, Zhengyang and Chen, Pei and Li, Zheng and Li, Xian and others},
  booktitle={Proceedings of the 63rd Annual Meeting of the Association for Computational Linguistics (Volume 1: Long Papers)},
  pages={6936--6949},
  year={2025}
}

@article{lin2025refrag,
  title={Refrag: Rethinking rag based decoding},
  author={Lin, Xiaoqiang and Ghosh, Aritra and Low, Bryan Kian Hsiang and Shrivastava, Anshumali and Mohan, Vijai},
  journal={arXiv preprint arXiv:2509.01092},
  year={2025}
}

\appendix
\section*{Appendix}

\section{Datasets Details}
\label{sec: appendix_dataset}
\begin{table}[h]
\centering
\scalebox{0.85}{
\setlength{\tabcolsep}{4pt}{
\begin{tabular}{lrrrr}
\toprule
 & TopiOCQA & QReCC & CORAL & INSCIT \\ 
\midrule
\#Conv. & 205 & 2,775 & 200 & 469 \\
\#Turns(Qry) & 2,514 & 16,451 & 2,153  & 2,767 \\
\#Collection & 25M & 54M & 0.2M & 49M \\
\#Avg. Qry & 12.9 & 5.3 & 10.8 & 5.9 \\
\#Min Qry & 5 & 2 & 8 & 2  \\
\#Max Qry & 25 & 12 & 19 & 7 \\
\#Avg. Psg & 9.0 & 1.6 & 2.4 & 1.6 \\
\bottomrule
\end{tabular}}}
\caption{Statistics of the four used datasets.}
\label{table: datasets}
\end{table}
The statistics of each dataset are presented in Table~\ref{table: datasets}. For the CORAL dataset, we only use the most challenging subset (level-d) with the longest conversations for the evaluation in our experimental scenario.
Additionally, we utilize the INSCIT dataset for evaluating mixed-initiative actions, as it is the only dataset annotated with the associated ground-truth information. 

\section{Baseline Details}
\label{sec: appendix_baseline}
We provide a more detailed introduction to the following baselines used for comparison:

\textbf{Supervised fine-tuning (SFT)}: A simplest training method with next-token prediction loss to generate the answer on the backbone language models.

\textbf{ChatQA}~\cite{liu2024chatqa}: A two-stage instruction fine-tuned model for conversational QA and RAG, with synthetic data generation and human-annotated high-quality data.

\textbf{UniConv}~\cite{mo2025uniconv}: A unified framework to improve the seamless consistency between conversational retrieval and its augmented response generation within a single LLM.

\textbf{REFRAG}~\cite{lin2025refrag}: An optimized architecture to compress only a small subset of retrieved documents that are directly related to the query for effective and efficient RAG. 

\textbf{EvoRAG}~\cite{cheng2025evorag}: A novel framework that dynamically maintains an evolving knowledge graph to model the logical relations among user queries, system responses, and the relevant passages across conversational turns.

\textbf{Search-R1}~\cite{jin2025search}: A new and effective paradigm for enhancing input quality of the LLMs by integrating in-context reasoning with dynamic search tool invocation when needed.

\textbf{ConvSearch-R1}~\cite{zhu2025convsearch}: A conversational query rewriting approach that eliminates dependency on external rewrite supervision based on the paradigm of Search-R1.

\textbf{ChatR1}~\cite{lupart2025chatr1}: An RL-based reasoning model for Conversational QA, which optimizes multi-turn retrieval and generation end-to-end with dynamic behavior learning.

\textbf{AgenticLM}~\cite{yang2025qwen3}: Leveraging the capacity of agentic LLMs to perform explicit reasoning in conversational scenarios.

\section{Instruction Template and AI Usage}
The full instruction template to guide our agentic conversational model based on different backbone LLMs is provided in Table~\ref{tab: prompt}. Besides, we only use AI assistants (e.g., ChatGPT and Claude) to test in our experimental scenarios, without the use of coding and writing.
\begin{table}[h]
\centering
    \begin{lstlisting}[language=PolicyPrompt]
You are a helpful assistant. Answer the question in multi-turn conversations. The historical context is provided inside <context> and </context>. 
- You must conduct reasoning inside <think> and </think> first every time you get new information. After reasoning, if you find you lack some knowledge, you should first rewrite the context-dependent question into a rewritten query by refering the historical context. 
- Then, you can call a search engine by <search> rewritten query + your answer </search>, and it will return the top searched results between <information> and </information>. You can search as many times as you want. 
- If you find no available answer could be provided, you can say you cannot find any information inside <answer> and </answer>, without detailed illustrations. For example, <answer> Sorry, I did not find any useful information. </answer>. 
- If you find the question is ambiguous to answer or want to engage the conversation, you can generate a clarification question inside <clarify> and </clarify>, without detailed illustrations. For example, <clarify> There are commercial and homemade substitutes available, which ones would you like to know about? </clarify>. 
- If you find no further external knowledge needed, you can directly provide the answer inside <answer> and </answer>, without detailed illustrations. For example, <answer> Beijing </answer>.

Context Begin: <context> {Historical Context} </context>.
Question: {Current User Query}
    \end{lstlisting}
  \caption{Instruction Template for the policy LLM.}
  \label{tab: prompt}
\end{table}

\end{document}